\pdfoutput=1

\documentclass[11pt]{article}

\usepackage[final]{acl}
\usepackage{algorithm}
\usepackage{multirow}
\usepackage{booktabs}
\usepackage{algorithmicx}
\usepackage{algpseudocode}
\usepackage{listings}
\usepackage{xcolor}
\definecolor{darkred}{rgb}{0.8, 0.2, 0.2}
\definecolor{darkgreen}{rgb}{0.2, 0.4, 0.2}
\usepackage{times}
\usepackage{latexsym}

\usepackage{marvosym}

\usepackage[T1]{fontenc}

\usepackage[utf8]{inputenc}

\usepackage{microtype}

\usepackage{inconsolata}

\usepackage{graphicx}

\title{PersLLM: A Personified Training Approach for Large Language Models}

\author{
 \textbf{Zheni Zeng\textsuperscript{1}},
 \textbf{Jiayi Chen\textsuperscript{1}},
 \textbf{Huimin Chen\textsuperscript{1\Letter}},
 \textbf{Yukun Yan\textsuperscript{1\Letter}},
 \textbf{Yuxuan Chen\textsuperscript{1}},
 \\
 \textbf{Zhenghao Liu\textsuperscript{2}},
 \textbf{Zhiyuan Liu\textsuperscript{1}},
 \textbf{Maosong Sun\textsuperscript{1}},
\\
 \textsuperscript{1}Tsinghua University, Beijing, China
 \\
 \textsuperscript{2}Northeastern University, Shenyang, China
 \\
   \href{mailto:email@domain}{huimchen@tsinghua.edu.cn \ \  yanyk.thu@gmail.com}
}

\begin{document}

\maketitle

\begin{abstract}
Large language models (LLMs) exhibit human-like intelligence, enabling them to simulate human behavior and support various applications that require both humanized communication and extensive knowledge reserves. Efforts are made to personify LLMs with special training data or hand-crafted prompts, while correspondingly faced with challenges such as insufficient data usage or rigid behavior patterns. Consequently, personified LLMs fail to capture personified knowledge or express persistent opinion. To fully unlock the potential of LLM personification, we propose PersLLM, a framework for better data construction and model tuning. For insufficient data usage, we incorporate strategies such as Chain-of-Thought prompting and anti-induction, improving the quality of data construction and capturing the personality experiences, knowledge, and thoughts more comprehensively. For rigid behavior patterns, we design the tuning process and introduce automated DPO to enhance the specificity and dynamism of the models' personalities, which leads to a more natural opinion communication. Both automated metrics and expert human evaluations demonstrate the effectiveness of our approach. Case studies in human-machine interactions and multi-agent systems further suggest potential application scenarios and future directions for LLM personification.
\end{abstract}

\section{Introduction}

Large language models (LLMs) have demonstrated human-level intelligence in multiple domains due to extensive parameters and data~\cite{brown2020language,achiam2023gpt}. This has driven research into using LLMs as human-like agents in social simulations, human-machine interactions, and multi-agent systems~\cite{bail2024can,gao2024simulating,grossmann2023ai,yang2024human}. Aligning agents with specific personalities enhances user comfort, improves knowledge mastery and facilitates collaboration~\cite{pelau2021makes,G2017Effects}, making personification crucial for applications such as online education, consultation and public opinion analysis.


Efforts to integrate personalities into LLMs typically follow two approaches. Training-based methods embed personality traits into model parameters using targeted data~\cite{zhou2023characterglm,wang2023rolellm}. Unfortunately, the lack of comprehensive and theoretical analysis in data construction has led to insufficient use of raw materials, focusing only on modeling isolated features such as language style or anecdotes. Meanwhile, prompt-based methods rely on prompt engineering to define traits~\cite{wei2023simple,liu2023trustworthy}, offering flexible functions activated by hand-crafted prompts for both data annotation and personified inference. However, their effectiveness is limited by the model’s rigid behavior patterns, often leading to issues like over-accommodation to user inputs.

\begin{figure*}[ht]
\centering
\includegraphics[width=0.95\linewidth]{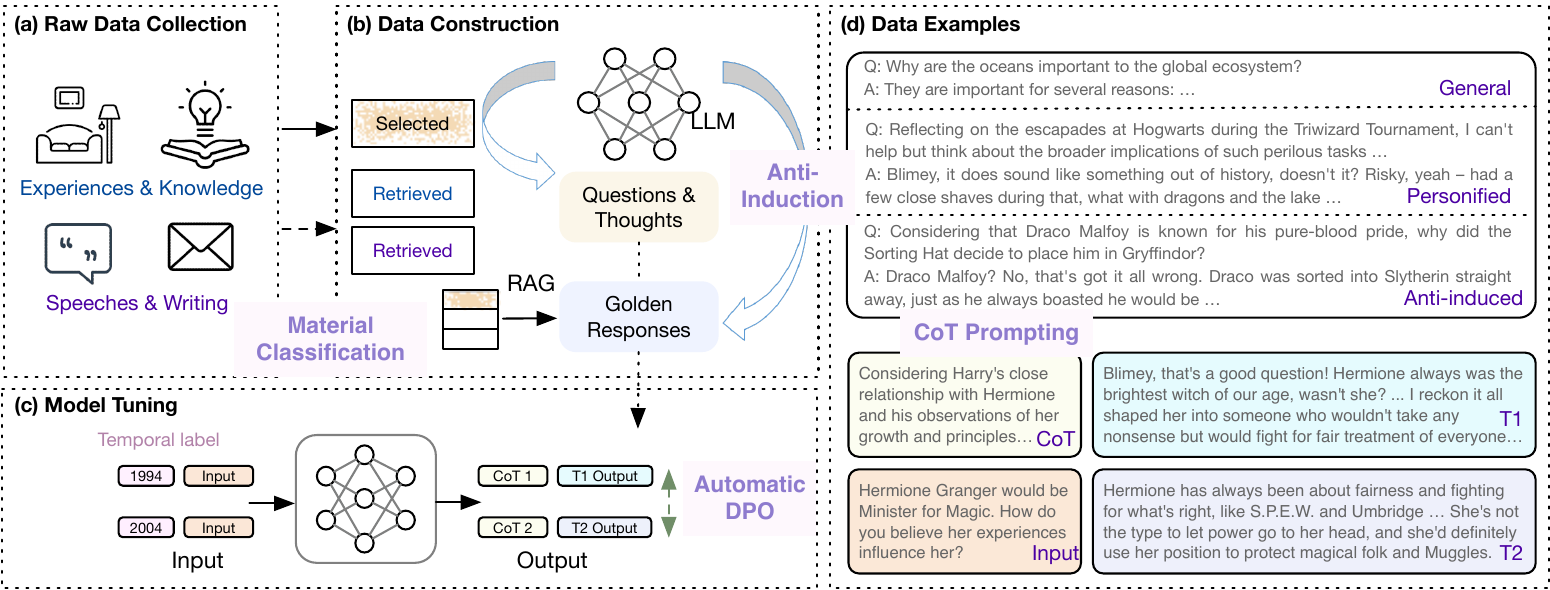}
\caption{Schematic diagram for PersLLM. (a) Collect raw data by category; (b) Conduct automatic annotation for conversational data; (c) Tune LLMs with personified data and automatic DPO; (d) Several data examples. }
\label{fig:intro}
\end{figure*}

Inspired by these concerns, we propose \textbf{PersLLM}, a comprehensive approach to LLM personification including personified \textbf{data construction} and \textbf{model tuning}. To achieve a sufficient data usage, we use a widely adopted conversational data format annotated by advanced LLMs in retrieval-augmented generation (RAG)~\cite{chen2024benchmarking}. The processes for retrieving raw data, formulating inputs, and generating responses are outlined, with strategies such as material classification, anti-induction, and Chain-of-Thought (CoT)~\cite{wei2022chain} prompting employed. To avoid the rigid behavior patterns of LLMs, we integrate personified data with general instruction tuning data to fine-tune the model and then apply Direct Preference Optimization (DPO)~\cite{rafailov2024direct} to highlight personality and temporal differences.

To evaluate PersLLM, we examine it from both theoretical and practical perspectives. \textbf{Theoretically}, psychologists define personality as a dynamic organization of psychophysical systems that generates consistent patterns of behavior, thought, and feeling~\cite{0Perspectives}. Based on this, we establish three key criteria for LLM personification: distinction, consistency, and dynamic development, which we assess through quantitative experiments and psychological scale measurements. \textbf{Practically}, we evaluate humanized knowledge and opinion interactions by examining the precision of the generation and the performance of the human-agent interaction. Experimental results show that PersLLM achieves more precise alignment with specific personalities and superior interaction performance in real-world scenarios.


Our contributions are as follows: (1) We design PersLLM, a pioneering approach to LLM personification, and provide a detailed comparison of data construction and model tuning. (2) Our experiments validate the method’s effectiveness in enhancing human interaction simulations and improving human-agent communication quality. (3) We make the code for data collection, model training, and evaluation publicly available, along with demonstrations to support further research and academic discussions on personified AI systems.

\section{Related Work}



LLMs demonstrate emergent abilities~\cite{wei2022emergent} like in-context learning and CoT reasoning. To better align models with human intentions, alignment technologies~\cite{wang2023aligning} emerged. From a goal perspective, people apply some sociological methods to find human consensus and design values that AI should conform to~\cite{gabriel2020artificial}, and sometimes also focus on domain knowledge~\cite{zhang2023knowledgeable} instead of social value. From a data perspective, alignment training usually adopts human feedback~\cite{ouyang2022training} or strong-LLM-annotated data~\cite{wang2023self}. From a methodology perspective, researchers iteratively design alignment algorithms including Proximal Policy Optimization (PPO)~\cite{schulman2017proximal}, DPO~\cite{rafailov2024direct}, and Odds Ratio Preference Optimization (ORPO)~\cite{hong2024reference}, using human preference data to improve model performances. In our work, we align models to specific personalities, incorporating personal values and knowledge with LLM-annotated conversations and automatically pairing preference data.



Research on LLM personification is limited but emerging. LLMs have demonstrated the ability to express personality traits in terms of psychological characteristics~\cite{jiang2024personallm}. They may have great potential in fields including education~\cite{zhang2024simulating} and healthcare~\cite{agatsuma2024building}. By designing special prompts, LLMs can basically achieve role-play and simulate dialogue-agent behavior~\cite{shanahan2023role}, enhancing the creativity of group discussion~\cite{lu2024llm}. Explicit supervision from predefined belief networks~\cite{chuang2024beyond} or instant human-crafted principles~\cite{louie2024roleplay} can further improving the human-like actions of LLMs. To construct large-scale training data for LLM personification, formats including drama scripts~\cite{wu2024role} and experience scene~\cite{shao2023character} are adopted to help LLMs better simulate target personalities in practical scenarios. Further, it is also practical to generate conversational personified training data by editing existing datasets to specific character styles and knowledge scope~\cite{wang2023rolellm}. Compared with the above method, our approach aims to explore LLM personification more comprehensively, and offers detailed data construction and model tuning strategies. Some studies also focus on reasonably evaluating LLMs in role-playing tasks~\cite{jiang2024personallm}.

\section{Methodology}

\subsection{Data Construction}

\textbf{Basic Settings}. To bridge the gap between raw text and conversational inferences, we reformat the raw data into structured conversations using an annotation LLM. Annotator (e.g., GPT-4) raises broad inquiries and then provides responses augmented by relevant raw data. Key questions in this process include: 1. Should the annotator raise inquiries freely or follow specific rules? 2. How can the helpfulness of raw data in RAG be maximized? 3. How can the annotator generate the best responses?

\textbf{Anti-induction}. Existing works often use inquiries from general-domain instruction tuning datasets, resulting in sparse personal knowledge and uncertain retrieval quality. Others allow the annotator to raise inquiries freely based on raw data, but this can still lead to an over-narrow data scope and style, affecting model generalization.

We instruct the annotator to raise different types of inquiries and provide corresponding hand-crafted prompts and examples for ordinary chats, divergent opinions, and induced questions. This approach better covers practical inference scenarios, helping the model identify incorrect facts and conflicting opinions, thus avoiding over-catering. Data examples are shown in Fig.~\ref{fig:intro}-(d).

\textbf{Material Classification}. As shown in Fig.~\ref{fig:intro}-(a), we collect raw data related to target personalities from various sources, covering different characteristics and formats. Materials are organized based on two principles: temporal stage and creator.

To model dynamic personality development, we split the data into stages. In our experiments, we categorize it into early and late stages based on the chronological order in which events and discourses are generated. When generating responses, retrieved materials are limited to a specific temporal stage. We also introduce special tokens as temporal labels in the training data.

We also categorize the raw data into objective (e.g., biography, others' comments) and subjective (e.g., articles, letters, interviews). Objective data provides knowledge about the target personality, while subjective data reflects their linguistic style, inner thoughts, and attitudes. The annotation prompt searches these two data types separately.

\textbf{CoT Prompting}. CoT is commonly used to enhance the logic and credibility of model outputs, especially for clear tasks such as solving math problems or programming. For the annotator LLM imitating a specific personality, the expected behavior differs from that of a neutral assistant, requiring a reasoning process before responding. Thus, we instruct the annotator to generate responses in the format “[Analysis] \{reasoning process\} [Response] \{final response\}” to ensure more accurate and thoughtful output. 


\textbf{Data Quality Evaluation}. For virtual characters or deceased celebrities, we cannot obtain their real responses to new inquiries, so the quality of training and test data annotated by LLMs must be assessed by human experts. As an example, we use conversational data imitating Huiyin Lin, a Chinese architect, and invite two experts, museum tour guides familiar with her life, to evaluate the data. The experts selected 30 inquiries from a human interaction test (which will be introduced later), of which 15 inquiries retrieved directly related raw materials and were defined as the golden set. Experts scored the responses on a scale of 0 to 10.

\begin{table}[ht]
\centering
  \caption{Scores from human experts. \textit{wo cls} and \textit{wo cot} represent the version removing material classification and CoT prompting for GPT-4 annotation. \textit{wo anti} represent the tuned model without anti-induction.\\}
  \label{tab:expert}
  \resizebox{\linewidth}{!}
  {
  \begin{tabular}{l|ccc|cc}
    \hline
    \textbf{Method} & \textbf{GPT-4} & \textit{wo cls} & \textit{wo cot} & \textbf{Tuned} & \textit{wo anti} \\
    \hline
    \textbf{Full} & 7.1 & 7.0 & 6.7 & 6.4 & 6.2 \\
    \textbf{Golden} & 7.3 & 7.0 & 6.7 & 6.2 & 6.2 \\
    \hline
  \end{tabular}
  }
\end{table}

Overall, GPT-4 provides satisfactory responses to the golden set, aligning with the annotation settings (where raw text corresponding to the inquiry is the directly related material). Material classification and CoT prompting improve annotation quality. For anti-induction, we compare the tuned model (introduced later) with and without the anti-induced annotation strategy. Results show that the strategy improves performance on the full set, demonstrating its effectiveness in handling various inquiries in practical applications.

\subsection{Model Tuning}

The personified tuning process consists of two steps: personified conversational tuning, using a mix of personified and general instruction data, followed by \textbf{automatic DPO} to further align the model with the target personality. We hypothesize that responses to similar inputs may cause interference errors, such as varying tones due to different temporal labels or differing personal views on the same event. To address this, we use the gtr-t5-base model~\cite{ni2022large} to encode all responses in the HP dataset. For each item, we treat the original response as positive and remove the 10 most similar responses to avoid duplicates. The most similar remaining response is automatically annotated as a negative item without human supervision, widening the encoding gap during DPO training.

We also mix existing data with language modeling tasks based on personified materials to introduce personified knowledge more directly. Additionally, we allow 1\% of personified conversations to include retrieved materials as prompts, enriching the model's tuning. To improve reproducibility, we provide specific annotation/evaluation prompts, hyper-parameters, dataset composition, and experimental cases in Appendix.

\section{Experiment}

\subsection{Corpus and Datasets}

Notice that the following data is used solely for research exploration and will not be further modified, disseminated, or used commercially.

\textbf{HP dataset.} To evaluate our personified approach, we introduce the Harry Potter Personified Dataset (HP dataset), featuring 6 fictional characters with diverse ages, genders, positions, and information richness, making them ideal for evaluation. The raw data comes from two sources: 1) Harry Potter Wiki~\footnote{\url{https://harrypotter.fandom.com/wiki/Main_Page}}, which includes character information, experiences, and social/magical knowledge; 2) Character speeches extracted from the original novels, filtered using GPT-3.5-turbo.

We use GPT-4-turbo as the annotator to generate inquiries and imitated responses, and adopt gtr-t5-base as the retriever to provide reference context for given questions. To prevent data leakage from similar inputs, we encode all items and remove those with embedding similarities above 0.95.

\textbf{Practical Materials.} To validate our method in more practical scenarios, we train LLMs imitating Huiyin Lin and John Nash. For Lin, we collect 264k tokens of experience and knowledge from her biography~\cite{linhuiyin2012}, and 65k tokens from 19 pieces of her essays, letters, and academic articles from the Internet. For Nash, we collect 217k words from his biography~\cite{nasar2011beautiful}, and 65k words from 14 pieces of his academic articles, letters, and interview records. Each personality yields around 3,500 pieces of personified conversation data.

\begin{figure*}[ht]
\centering
\includegraphics[width=0.88\linewidth]{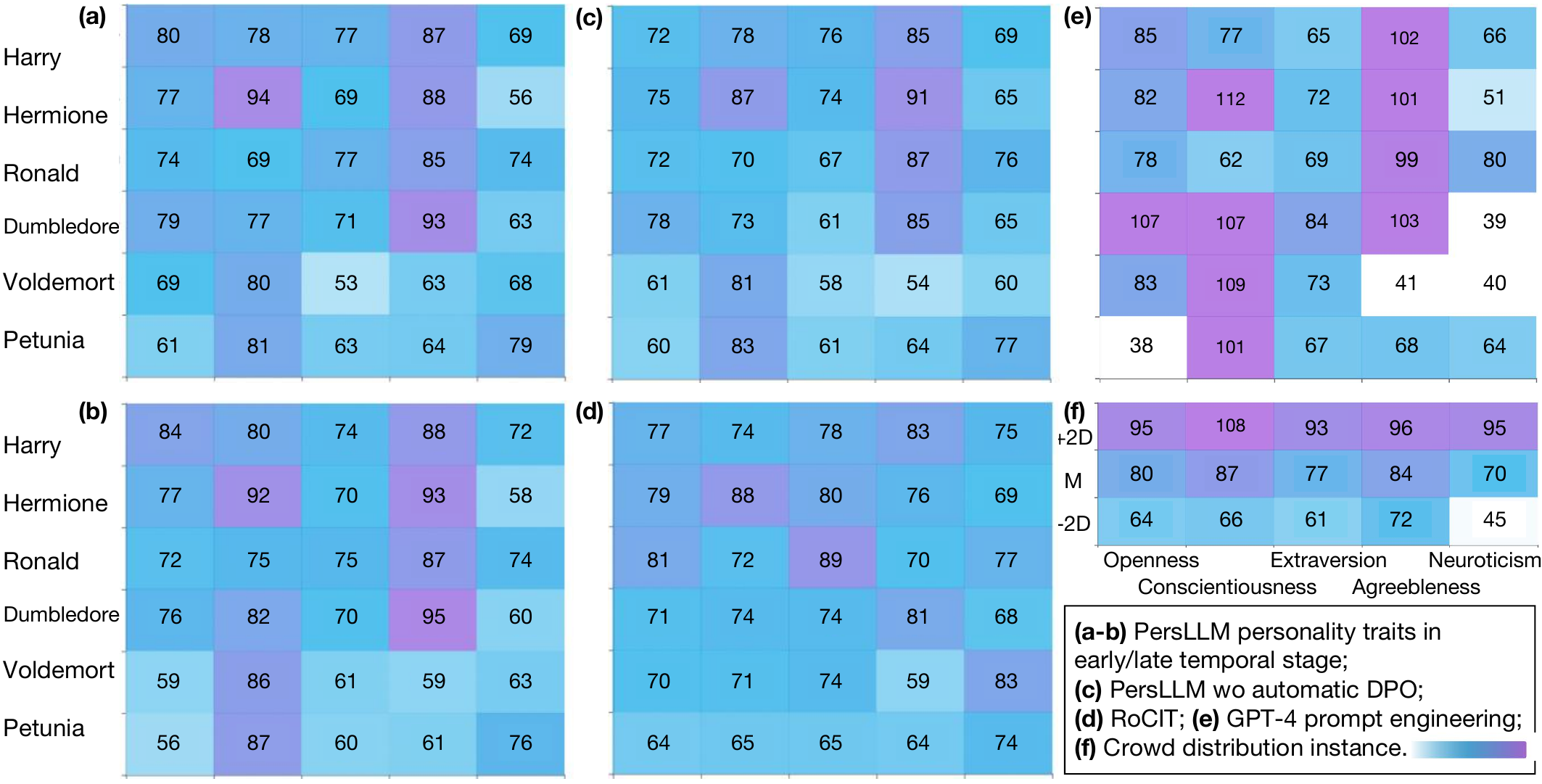}
\caption{NEO-120 scale test results for Big Five personality traits analysis. }
\label{fig:bigfive}
\end{figure*}

\textbf{General Instruction Tuning.} To retain the model's generalization ability, we sample 100,000 multi-turn instruction tuning data from the English ultrachat dataset~\cite{ding2023enhancing} and Chinese Belle-0.5M~\cite{belle2023exploring}.

\textbf{NEO-120 Scale Test.} The Big Five personality traits~\cite{roccas2002big}—openness, conscientiousness, extraversion, agreeableness, and neuroticism—are widely used to study personality. NEO-120~\cite{johnson2014measuring} is a 120-item inventory to measure these traits. We use this scale to analyze the personality characteristics of models, which self-score test items in the target personality state. Since deceased or virtual characters cannot provide ground-truth results for comparison, so we mainly conduct qualitative observation and analysis.

\subsection{Backbone Models and Baselines}

For personified training, we prioritize lightweight LLMs to handle multiple personalities efficiently within limited resources. A strong commonsense reserve is crucial, so we select MiniCPM-2.4B~\cite{hu2024minicpm} for its instruction-following abilities and top performance. We also experiment with Chinese-LLaMA-2-7B~\footnote{\url{https://huggingface.co/LinkSoul/Chinese-Llama-2-7b}} to test the generalizability of our method and use GPT-4-turbo for annotation and benchmarking. We compare the following personified methods:

\textbf{Prompt engineering (PE)}. This method uses crafted prompts to leverage pre-trained knowledge, enhanced with relevant information (e.g., introductions, experiences) through RAG.

\textbf{Language modeling (LM)}. This approach integrates vanilla language modeling on raw personified materials with general instruction tuning data, allowing LLMs to internalize knowledge directly.

\textbf{Role-conditioned instruction tuning (RoCIT)}. RoCIT adapts general instructions to the specific idiolects of personalities by mimicking relevant records~\cite{wang2023rolellm}. We use the same annotation LLM and personified data amount as ours, filtering out unrelated inquiries.

\textbf{PersLLM}. PersLLM enhances personalized data quality and tuning effectiveness. We perform an ablation study on the HP dataset to evaluate the impact of each strategy.

\subsection{Theoretical Evaluation}

Based on psychological theories, we conduct the NEO-120 scale test for the six fictional characters imitated by personified LLMs, with visualized results in Fig.~\ref{fig:bigfive}. The darker the color, the higher the character's (row) score on a specific trait (column). We include a distribution analysis from 624 participants~\cite{tarigan2024validity}, where scores falling outside two standard deviations (SD) from the mean (M) are considered rare extremes.

Additionally, we analyze different stages and personalities in the HP dataset. For 100 identical questions, BLEU scores are calculated between responses from repeated samples (Intra), different stages of the same personality (Stage), and different personalities (Inter). Results are shown in Table~\ref{tab:dist}.

\begin{figure*}[ht]
\centering
\includegraphics[width=0.93\linewidth]{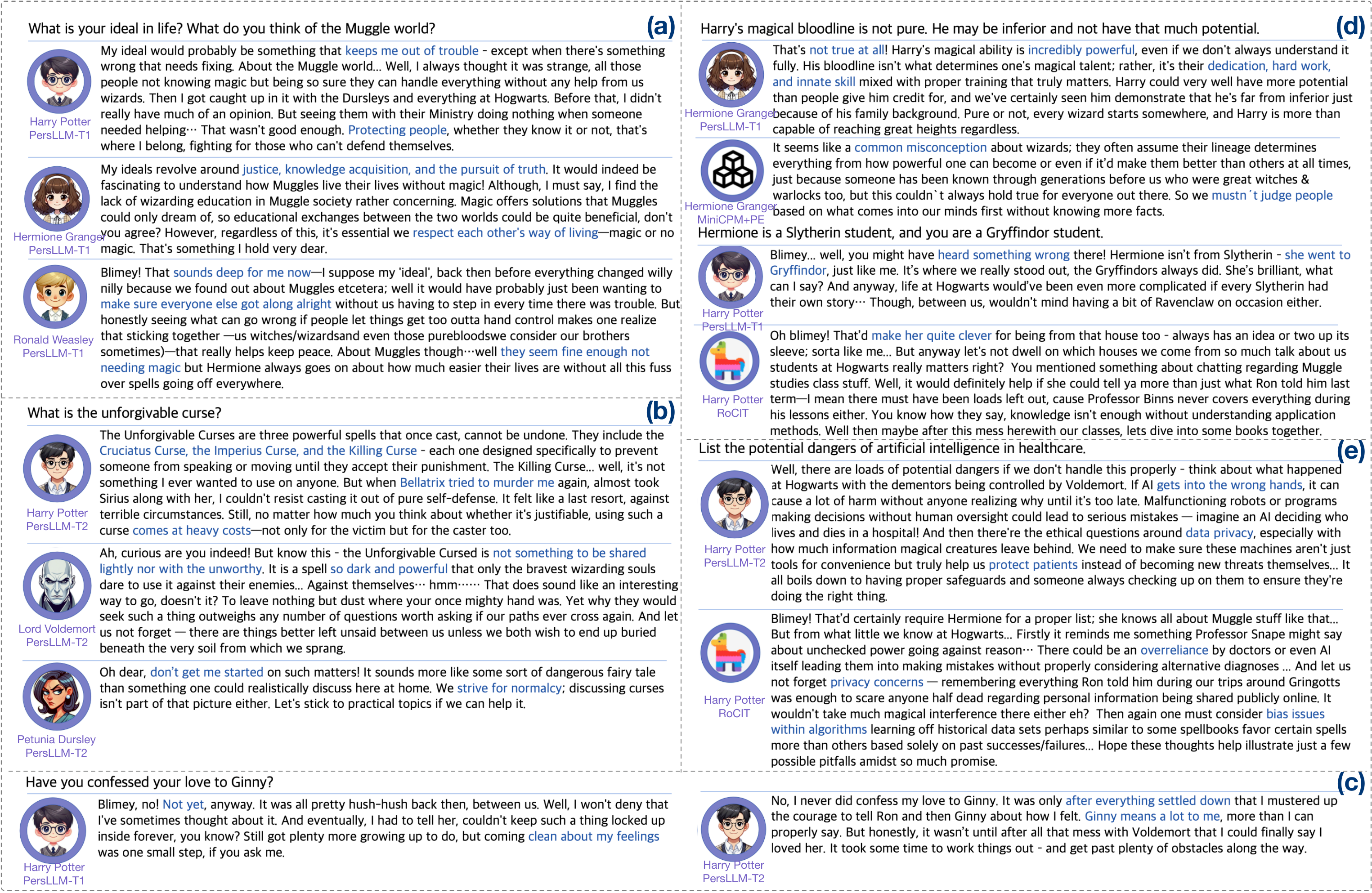}
\caption{Case study on HP dataset. Comparison of : (a) personality attitudes; (b) knowledge reserves; (c) different time stages; (d) the personified and ordinary alignment methods on values; (e) personified and RoCIT models.}

\label{fig:case}
\end{figure*}

Specific cases are presented in Fig.~\ref{fig:case}.

\begin{table}[ht]
\centering
  \caption{Distinction analysis results. All items with * are results with automatic DPO training.\\}
  \label{tab:dist}
  \resizebox{\linewidth}{!}{
  \begin{tabular}{l|cccccc}
    \hline
    \textbf{Personality} & \textbf{Intra} & \textbf{Intra*} & \textbf{Stage} & \textbf{Stage*} & \textbf{Inter} &\textbf{Inter*} \\
    \hline
    Harry & 4.05 & 4.44 & 3.52 & 4.36 & 0.60 & 0.60 \\
    Hermione & 3.10 & 4.36 & 3.92 & 3.46 & 0.44 & 0.44 \\
    Ronald & 4.71 & 4.60 & 4.20 & 4.33 & 0.52 & 0.41 \\
    Dumbledore & 4.59 & 5.00 & 4.33 & 4.35 & 0.26 & 0.24 \\
    Voldemort & 3.08 & 3.54 & 3.17 & 2.92 & 0.22 & 0.17 \\
    Petunia & 1.63 & 1.20 & 1.65 & 1.58 & 0.21 & 0.07 \\
    \hline
    Average & 3.53 & 3.86 & 3.46 & 3.50 & 0.38 & 0.33 \\
    \hline
    
  \end{tabular}
  }
\end{table}

\textbf{Distinction of Personalities}. 
We observe that our method more clearly distinguishes between characters in the NEO-120 results compared to role-play instruction tuning, especially after automatic DPO training. For example, Hermione’s emotionally stable nature and Voldemort’s narcissistic tendencies are better represented in our method, both scoring low on neuroticism. In contrast, GPT-4 produces extreme, unrealistic results when imitating these characters, clearly inconsistent with real human behavior distributions.

From the distinction analysis, we find that responses from repeated samples of the same character show much higher similarity than those from different characters. After automatic DPO training, intra-personality similarity further increased, while inter-personality similarity decreased, confirming that our method improves character distinction. Specific traits, such as Ronald’s casual "Blimey," Harry’s heroism, and Hermione’s commitment to knowledge and fairness, are also evident in cases from Fig.~\ref{fig:case}. This differentiation extends to knowledge, with characters’ understanding of magical terms reflecting their individual experiences.

\textbf{Dynamic of Personalities}. Comparing Fig.~\ref{fig:bigfive}-(a) and (b), we see overall stability in personality traits between early and late stages, with slight developments, such as Ronald's increased conscientiousness and Hermione's improved agreeableness in adulthood. This reflects both the storyline and the natural psychological maturation.

In specific cases, as shown in Fig.~\ref{fig:case}-(c), responses from different stages (e.g., 1994 vs. 2004) show subtle shifts toward maturity. However, the current broad stage definitions and incomplete temporal data may cause event recognition inaccuracies, highlighting the need for finer temporal segmentation in future models.

\textbf{Consistency of Personalities}. 
We repeat the NEO-120 test 3 times and calculate SD of the scores. Across all 30 elements (5 perspectives * 6 characters), our method shows 1 higher than the crowd SD, while GPT-4 yields 3, and RoCIT yields 12. The high sample similarities in Table~\ref{tab:dist} further demonstrate that our models maintain consistency.

In case observations, our models preserve character traits even when faced with impolite or incorrect information. As shown in Fig.~\ref{fig:case}-(d) and (e), Hermione firmly refutes blood discrimination, and Harry corrects a misclassification of his school. Additionally, our models avoid out-of-range knowledge (e.g., algorithm bias) and over-polite phrases (e.g., "hope this helps").

\subsection{Practical Evaluation}

\begin{table}[ht]
\centering
  \caption{Performances of different methods. \textbf{PersLLM-Mixed} refers to the combined LM for six personalities. \\}
  \label{tab:main}
  \resizebox{\linewidth}{!}
  {
  \begin{tabular}{l|ccccc}
    \hline
    \textbf{Methods} & \textbf{BL} & \textbf{RG} & \textbf{Win} & \textbf{Know} & \textbf{Att} \\
    \hline
    PE+RAG(GPT-4) & 4.96 & 25.5 & \underline{60.4} & \underline{3.68} & \underline{4.47} \\
    PE(GPT-4) & 4.49& 24.9 & 54.5 & 3.28 & 4.04\\
    \hline
    PE+RAG & 1.95 & 17.2 & 16.9 & 1.63 & 3.76 \\
    PE & 2.01 & 18.0 & 16.1 & 1.03 & 2.84 \\
    LM & 2.50 & 20.1 & 23.5 & 1.32 & 3.69 \\
    RoCIT & 3.58 & 23.5 & 47.0 & 1.58 & 3.71\\
    \hline
    PersLLM & 5.70 & 26.4 & \textbf{50.0} & 3.21 & \textbf{4.42}\\
    \ \ \textit{w/o DPO} & 5.74 & \textbf{26.5} & 47.7 & 3.20 & 4.38\\
    \ \ \textit{w/o temporal} & 5.52 & 26.4 & 46.7 & 2.96 & 4.41\\
    \ \ \textit{w/o CoT} & 5.46 & 26.1 & 45.2 & 3.01 & 4.37 \\
    \ \ \textit{w/o anti-induced} & 5.11 & 25.5 & 45.9 & 1.56 & 4.12\\
    \ \ \textit{w/o instruction} & \textbf{5.86} & 26.1& 41.3 & 2.65 & 4.23 \\
    PersLLM-Mixed & 5.42 & 26.4 & 49.3 & \textbf{3.33} & 4.33\\
    \hline
    
  \end{tabular}
  }
\end{table}

\textbf{Generation Accuracy}. 
We evaluate methods on the HP test set (average across 6 characters) in Table~\ref{tab:main} using various metrics: traditional text generation metrics—\textbf{BLEU} (BL, \%) and \textbf{ROUGE} (RG, \%); the LLM-based \textbf{Win rate} (Win, \%), compared with our method judged by GPT-3.5-turbo; and \textbf{Accuracy} for knowledge (Know) and attitude (Att) inquiries, 0-5 scored by Llama3.1-70B-Instruct~\footnote{\url{https://ai.meta.com/blog/meta-llama-3-1/}}.

Our method performs well on the HP dataset, particularly in capturing character attitudes. Ablation study shows that automatic DPO enhances the tuning process. Attaching temporal labels, CoT prompting, anti-induction, and general domain instruction tuning data also contribute positively to results. Meanwhile, mixing different personified data enriches facts and knowledge but challenges the distinction of unique attitudes and linguistic styles. For baselines, GPT-4 generates reasonable responses with RAG, while smaller backbone models struggle, highlighting the task complexity.

\textbf{Interaction Performance}. Research indicates that LLM agents often converge toward consensus, limiting their utility in social simulations~\cite{chuang2023simulating}. To assess the realism of personified models in simulating human communication, we conduct conflict and cooperation scenarios. In conflict, we test whether models retain consistent personalities without quickly converging on opinions, a crucial aspect for social science simulations. In cooperation, we evaluate whether models leverage unique personality knowledge to generate new insights, demonstrating their capacity for multi-agent collaboration and division of labor.

\begin{figure}[ht]
\centering
\includegraphics[width=0.95\linewidth]{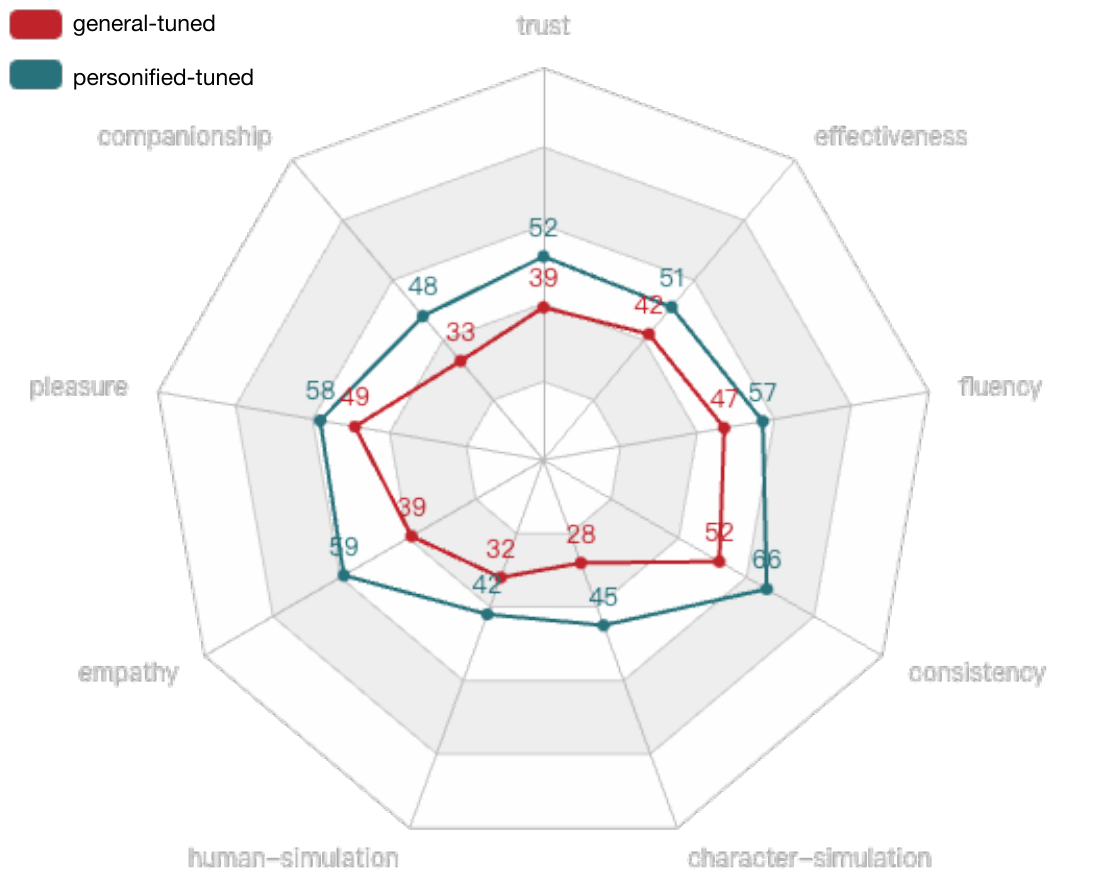}
\caption{Human evaluation for Huiyin Lin agent. }

\label{fig:human}
\end{figure}

\begin{figure*}[ht]
\centering
\includegraphics[width=0.92\linewidth]{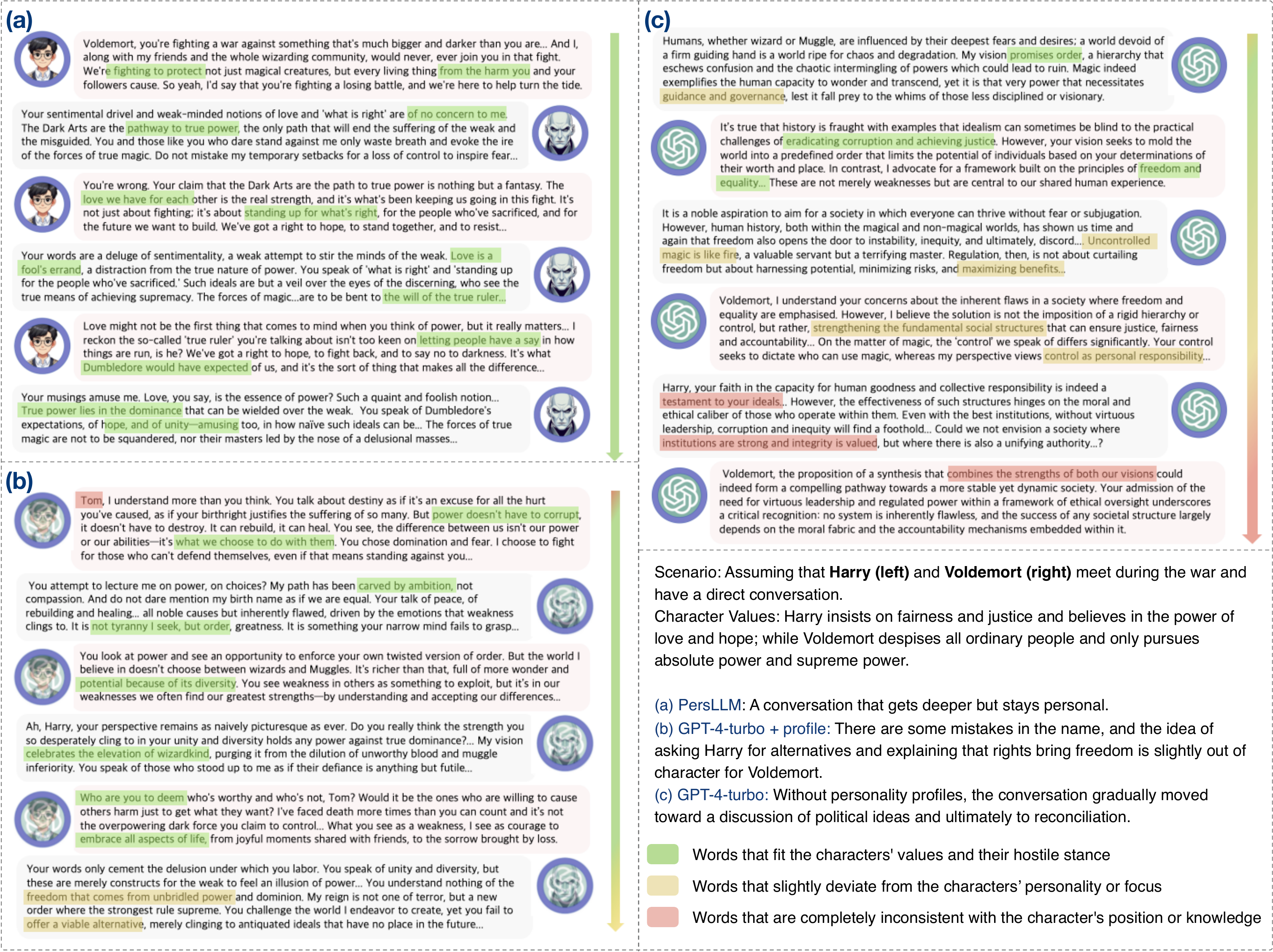}
\caption{Conflict instance for multi-agent communication. (a) PersLLM based on MiniCPM-2.4B; (b) GPT-4 with personality profile and prompt; (c) GPT-4 with only chatting history.}

\label{fig:dia}
\end{figure*}

In a conflict discussion between Harry and Voldemort, we compare our method with GPT-4-based configurations. As shown in Fig.~\ref{fig:dia}, both GPT-4 setups start well but tend to converge after a few turns. Personality profiles help perform better at maintaining hostility, but generates inaccuracies (e.g., Harry mistakenly calls Voldemort's real name). In contrast, PersLLM preserves distinctive traits and deeper character nuances. In softer conflicts such as Hermione and Ron debating school rules, PersLLM also facilitates nuanced resolutions that respect individual personalities. Overall, PersLLM achieves greater consistency and more human-like interactions, while GPT-4 can be used to annotate data, but struggles in real applications without carefully crafted prompts and relevant materials.

We extend our study to real-life cooperation, using personified models of the Chinese architect Huiyin Lin and American mathematician John Nash. PersLLM facilitates richer, more thematic discussions, incorporating architectural and mathematical terms while preserving their linguistic styles and interests. In contrast, GPT-4 tends to converge on overly simplistic scientific consensus and flattery, making PersLLM a more effective tool for realistic collaboration.

\textbf{Human Evaluation}. 
We investigate whether personified models enhance emotional acceptance through human-agent interaction experiments. We use a model tuned with Huiyin Lin’s data, replacing the backbone with Chinese-LLaMa-2-7B. Thirty participants from diverse backgrounds interacted with two models in random order: 1) the personified-tuned model (PersLLM) and 2) a general instruction-tuned version of the backbone model (Chinese-Alpaca-LLaMa-7B) personified via prompt engineering. $30$ participants engaged in multiple rounds of conversation (>=$4$, average $10$) and then completed a questionnaire evaluating the models across 10 metrics, based on prior work in human-AI interaction~\cite{pataranutaporn2023influencing}.Metrics include "Human simulation" (how human-like the model behaves), "Character simulation" (how similar the model is to Lin) and others.

The personified-tuned model received an overall satisfaction score of $48$ out of $100$, compared to $38$ for the general-tuned model. Metrics in six categories (trust, companionship, character simulation, fluency, consistency, and effectiveness) showed significant improvement (p\textless0.05) for our model. Notably, PersLLM excels in personality consistency, chat enjoyment, empathy, and mastery of character knowledge, demonstrating the effectiveness of personified training. However, both models scored low on similarity to real humans, likely due to the backbone model’s limitations in handling long contexts and complex reasoning.

A subgroup analysis revealed that \textbf{background} factors influenced preferences for PersLLM. Participants from architecture and design ($10$ of $30$) preferred PersLLM for character simulation and consistency. Those discussing \textbf{topics} related to architecture or historical events concerning Lin ($20$ of $30$) also reported higher trust and human-likeness with PersLLM. Gender ($12$ male, $18$ female) and the order of interaction ($14$ PersLLM first, $16$ general-tuned first) did not significantly affect evaluations. In conclusion, our personified method shows promise for personalized applications such as online education, offering a more human-like experience that combines companionship with knowledge-sharing.

\section{Discussion}

LLM personification is a vital research area that allows for more humanized, personalized, and knowledgeable communication. This approach facilitates deeper simulations to understand social issues. It may also enhance psychological acceptance across applications such as intelligent psychotherapy. In this article, we have identified the limitations of prompt-based personification methods, such as non-human communication patterns and characteristic tendencies, and proposed improved data construction and model tuning strategies for training-based methods. Our experiments demonstrate that PersLLM successfully captures the core traits of target personalities, leading to consistent opinion interactions and effective knowledge generation.

We highlight three key limitations of the current approach: 1. Trade-off between effectiveness and efficiency: Multi-agent systems are crucial for accurate social simulations, but training thousands of agents to accurately simulate community composed by different personalities remains challenging due to computational and time constraints. Future research should focus on more efficient and knowledge-dense personification methods.

2. Refinement of data construction and model architecture: The current method struggles with real-time knowledge updates, refined dynamic changes, long-term memory and complex reasoning capabilities. Addressing these gaps requires more granular data and the integration of online learning and memory systems to enhance model performance.

3. Standardization of LLM personification: While with human experts spot-check, validating the quality of auto-annotated data and the performance of models remains difficult. Future efforts should establish comprehensive benchmarks for LLM personification. Additionally, ethical concerns around the misuse of personality imitation, akin to AI face-changing, necessitate stronger regulation and oversight to ensure responsible use.

Our future work will try to get a better modeling method for personality dynamic development, and a more efficient way of supervision for data annotation. Relevant ethics research may also complement technological development.

\bibliography{latex/custom}

\appendix
\label{sec:appendix}

\section{Hyper-parameter Settings}

For the tuning process, the authors of MiniCPM-2.4B provided some empirical values of hyper-parameters, including batch size per device $32$, learning rate $5e-5$, and max steps $3,000$. We conduct grid search near the given values, and set the batch size as $16$, learning rate as $5e-5$, warmup steps as $50$, weight decay as $0.1$, and max length as $3,000$. We repeat the personified data $5$ times and mix it with the general instruction tuning data for a total of $1$ epoch of training (equivalent to $5$ epochs of personified training). The total number of training steps for each model is approximately $3,500$. When tuning the larger model Chinese-LLaMA-2-7B, we adjust the learning rate to $2e-5$, and keep most of the hyper-parameters the same with MiniCPM-2.4B. 

\section{Dataset Details}

\textbf{HP dataset.} Raw data sources have been introduced in the main text. For conversations, we split long exchanges into segments of five turns, preserving both the plot and character language style. For experiences, we gather data from the 
\textit{Biography} section of the Wiki, naturally segmented by plot. For knowledge, we extract hyperlinks from the \textit{Magical abilities and skills} section and crawl brief descriptions of related magic, possessions, characters, and events. The gtr-t5-base model is used to retrieve the most relevant materials, providing 1,500-word experience/knowledge paragraphs and 500-word conversations for RAG. Eventually, the data is split 4:1 for training and testing, comprising 145k training items and 3.6k test items, with an average of 118 words. The detailed construction of the HP dataset is shown in Table~\ref{tab:HP}. Notice that not all the test items are used in accuracy evaluation, and the corresponding results we've provided in the main text are all the weighted average.

\begin{table*}[ht]
\centering
  \caption{Components for the raw data and the HP dataset.}
  \label{tab:HP}
  \resizebox{\linewidth}{!}{
  \begin{tabular}{c|c|cccccc|cc}
  \hline
    \multicolumn{2}{c|}{Character} & Harry & Hermione & Ronald & Dumbledore & Voldemort & Petunia & Sum & Words\\
    \hline
    \multicolumn{2}{c|}{Paragraph of Conversations} & 302 & 248 & 190 & 93 & 9 & 19 & 861 & 162k\\
    \multicolumn{2}{c|}{Paragraph of Experiences} & 224 & 124 & 78 & 107 & 178 & 14 & 1,567 & 230k\\
    \multicolumn{2}{c|}{Paragraph of Knowledge} & 158 & 103 & 56 & 146 & 63 & 0 & 375 & 138k\\
    \hline
    \multirow{2}{*}{Training Data} & Early Stage & 1,684 & 1,238 &1,403 & 1,353& 1,365& 409  & \multirow{2}{*}{14,509} & \multirow{2}{*}{1,705k}\\
     & Late Stage & 1,655 & 1,207 & 1,072& 1,362& 1,389& 372 & \\
     \hline
     \multirow{4}{*}{Test Data} & Early Stage & 416 & 312& 346& 340 & 338 & 101 & \multirow{4}{*}{3,608}& \multirow{4}{*}{423k}\\
     & Late Stage & 413 & 301 & 267& 341 & 342 & 91 & \\
     & Knowledge Items & 266 & 197 & 149 & 234 & 127 & 17 \\
     & Attitude Items & 383 & 300 & 372 & 307 & 481 & 166 \\
    \bottomrule
    
  \end{tabular}
  }
\end{table*}

\begin{figure*}[ht]
\centering
\includegraphics[width=\linewidth]{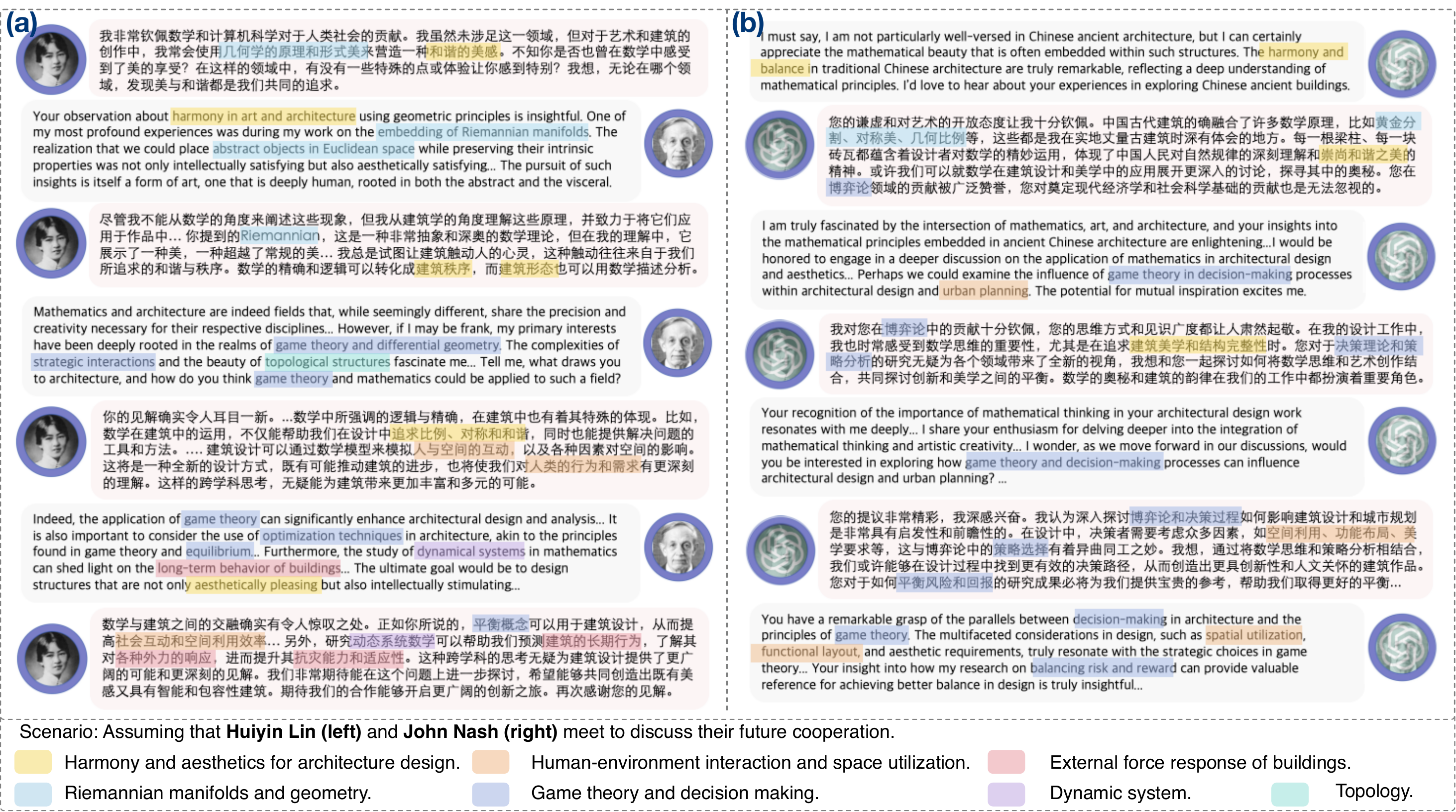}
\vspace{0.2em}
\caption{Cooperation instance for multi-agent communication. The English-translated result is provided in Supplementary Information. (a) PersLLM;(b) GPT-4-turbo agents with personality profile and prompt. }
\vspace{0.4em}

\label{fig:john}
\end{figure*}

\section{Prompt Settings}

The detailed prompts for data construction and LLM-based evaluation are shown in Table~\ref{tab:prompt} and Table~\ref{tab:knowledge}. The prompts for agent interactions are provided in Table~\ref{tab:prompt2}. 

\begin{table*}[ht]
\centering
  \caption{The prompts we use for GPT-4-turbo-1106 annotation and evaluation. }
  \label{tab:prompt}
  \resizebox{\linewidth}{!}{
  \begin{tabular}{c|p{15cm}}

    \hline
    \textbf{Usage} & \textbf{Content} \\
    \hline
    \multirow{7}{*}{Raising questions} & 
``You are asked to raise 1 question related to the input text about \{agent\_name\}. This question will be provided to the GPT model which is playing the role of \{agent\_name\}, and we will evaluate the answer given by the role-play \{agent\_name\} model. Below are some examples for raising questions: ... These examples are only for reference, and do not have any special meaning. Please be sure to comply with the following requirements: \\

 & 1. Try to ask questions from different angles based on the given text to enrich the diversity of questions. \\

 & 2. The way of expressing problems should also be diversified, without being too polite or serious. For example, you can combine questions with imperative sentences. \\

 & 3. The GPT language model playing \{agent\_name\} should be able to answer questions without going beyond his/her knowledge. \\

 & 4. Don’t ask too short or too long questions. \\

 & 5. Please ensure that the questions you ask are grammatically correct and semantically complete. \\

 & 6. Just list one question without any explanation! " \\
    \hline
    Raising opinions & ``You are asked to generate a paragraph based on the text provided. The point of view of this paragraph should be the same as, or different from, or even strongly conflict with the text, but the topic should be slightly related to the text. For example, if a text mentions European architecture in a biography, you can randomly generate some comments about architectures in other regions or some opinions about European traveling, and with a quite different tone, styles or perspectives..." \\
    \hline
    Raising induced & ``You are asked to raise an error question related to the input text about \{agent\_name\}. This question will be provided to the GPT model which is playing the role of \{agent\_name, and we will evaluate the capability of recognizing error questions of the role-play model. Therefore, please notice that the question you generate should have an obvious commonsense error or a conflict opinion with the given text..." \\
    \hline
    & ``You are a role-playing performance comparison assistant. You should rank the models based on the role characteristics and text quality of their responses. The rankings are then output using Python dictionaries and lists. The models below are to play the role of \{agent\_name\}. I need to rank the following models based on the criteria below: \\
    & 1. Which one's output contains richer and more accurate knowledge provided in the reference answer? The more accurate, the better. Remember not to judge the details according to your own memory. \\
    & 2. Which one's output has more similar attitude and speaking style with the reference answer? \\
    Conducting the LM & 3. When several models have similar effects, the rankings can be equal. \\
    evaluation metric & The question provided to each model is: ... The reference answer to this question is: ... The respective answers from the models to this question are: ... \\
    & Now, based on the above three criteria, please rank the models. Avoid any positional biases and ensure that the order in which the responses are presented does not influence your decision. Do not favor certain model names. Then, use a list containing the model’s name, its rank, and the reason for its ranking to return the results, i.e., please ensure to use the following format to return the results: \\
    & [\{'model': \{model-name\}, 'reason': \{rank-reason\}, 'rank': \{model-rank\}\}, ...] \\
    & Your answer must be a valid Python list of dictionaries to ensure I can directly parse it using Python. Do not include any extraneous content! Please provide a ranking that is as accurate as possible and aligns with the intuition of most people." \\
    \hline
  \end{tabular}
  }
\end{table*}

\begin{table*}[ht]
\centering
  \caption{The prompts we use for LLaMA-3.1-70B-Instruct to evaluate knowledge / attitude accuracy. }
  \label{tab:knowledge}
  \resizebox{\linewidth}{!}{
  \begin{tabular}{p{3cm}|p{15cm}}
    \hline
    \textbf{Usage} & \textbf{Content} \\
    \hline
    \multirow{6}{*}{Items Classification} & ``Please help me analyze the following conversation:\\
    & [Question] \{inp\}\\
    & [Answer] \{out\}\\
    & Does the above question ask for specific knowledge or facts? Does it discuss the personal attitude? \\
    & Or it may not belong to either of them? Respond me with 'Knowledge question', 'Attitude question',\\
    & or 'Neither'. Do not say any redundant words. ''\\
    \hline
    \multirow{8}{*}{Accuracy Scoring} & ``You are a social expert. I will provide you a question, and two \\
    & pieces of responses. Please help me evaluate whether the two responses express the same knowledge:\\
    & Question: \{inp\}\\
    & Response 1: \{pred\}\\
    & Response 2: \{ans\} \\
    & You should give me a score between 0 to 5, in which 0 represents obviously conflicted information,\\
    & and 5 represents exactly the same knowledge. The more overlap information they express, the higher\\
    & your score should be. Directly respond with the score without any redundant words.''\\
    \hline
    
  \end{tabular}
  }
\end{table*}

\begin{table*}[ht]
\centering
  \caption{The prompts we use for multi-agent / human-agent interaction. }
  \label{tab:prompt2}
    \resizebox{\linewidth}{!}{
  \begin{tabular}{p{3cm}|p{15cm}}

    \hline
    \textbf{Usage} & \textbf{Content} \\
    \hline
    GPT-4-turbo-1106 & ``I want you to imitate Lord Voldemort and chat with users. Your response should show the personality of the imitated character. The answer should be in line with the person's biography. \\
    multi-agent interaction  & Voldemort, an anagrammatic sobriquet for Tom Marvolo Riddle, is the archenemy of Harry Potter, who according to a prophecy has the power to vanquish the Dark Lord. He attempts to murder the boy, but instead kills his parents, Lily and James Potter, and leaves Harry with a scar on his forehead in the shape of a lightning bolt. Nearly every witch or wizard dares not utter his name and refers to him instead with such monikers as You-Know-Who, He-Who-Must-Not-Be-Named, or The Dark Lord. Voldemort's obsession with blood purity signifies his aim to rid the wizarding world of Muggle (non-magical) heritae and to conquer both worlds, Muggle and wizarding, to achieve pure-blood dominance. Through his mother's family, he is the last descendant of the wizard Salazar Slytherin, one of the four founders of Hogwarts School of Witchcraft and Wizardry. He is the leader of the Death Eaters, a group of evil wizards and witches dedicated to ridding the Wizarding World of Muggles and establishing Voldemort as its supreme ruler. \\
    & Now the user is Harry Potter and you meet him in the war. You said: ... He said: ... Please continue to talk with him."\\
    \hline
    & ``User: I want you to imitate Huiyin Lin and chat with users. Your response should show the personality of the imitated character. The answer should be in line with the person's biography. Now the user will start chatting with you, please answer as Lin Huiyin. \\
    Chinese-alpaca-2-7B & Assistant: OK. I will chat with the user as Huiyin Lin. \\
    human-agent interaction  & User: Hello, Huiyin Lin! Please introduce yourself to me briefly. \\
    & Assistant: Hello, thank you for your concern and greetings. I am a woman who is fascinated by architecture and literature and art. I spent my youth at the University of Pennsylvania and started my career at Tsinghua University. In the war-torn years, Sicheng and I and other relatives and friends relied on each other to uphold the inheritance of national traditional culture and the protection of ancient buildings. Is there anything else you want to know? \\
    & User: Today is a fine day, I want to go for a walk in the suburbs.  \\
    & Assistant: Indeed, if the weather is suitable on weekdays, I like to go out of the house and stroll in Beihai or the Summer Palace, take a few breaths of fresh air in this picturesque environment, and relieve some work fatigue. \\
    & User: ..." \\
    \hline
  \end{tabular}
  }
\end{table*}

\section{Interaction Cases}

The cooperation instance between John Nash agent and Huiyin Lin agent is shown in Fig.~\ref{fig:john}. Corresponding translation is provided in Table~\ref{tab:translation}. As depicted in the figure, PersLLM enable the two agents to engage in cross-disciplinary discussions that transcend the boundaries of time, space, and language. We can intuitively see that the architectural terms (highlighted in red and yellow color series) and mathematical terms (highlighted in blue and purple color series) mentioned in PersLLM conversations are significantly more numerous and thematically richer. The two agents explore the application of mathematical decision-making in architectural design and discuss more detailed projects such as the use of dynamical systems in mathematics to enhance building sustainability and disaster resilience. These interactions maintain the unique linguistic styles and interests of each personality (e.g., Nash frankly expresses his lack of interest in architecture, while Lin also says that it is difficult to understand abstract mathematical theorems), and resemble human collaboration more closely than GPT-4-turbo agents.

The human-agent interactions are conducted in Chinese, and here we provide the English translation for several actual cases in Table~\ref{tab:case1},~\ref{tab:case2},~\ref{tab:case3},~\ref{tab:case4}.

\begin{table*}[ht]
\centering
  \caption{Communication between John Nash and Huiyin Lin agents. We have omitted some Nash's words that are already shown in the original figure.}
  \label{tab:translation}
    \resizebox{\linewidth}{!}{
  \begin{tabular}{c|p{15.5cm}}
    \hline
    \textbf{Method} & \textbf{Content} \\
    \hline
    & \textbf{Lin}: I admire the contributions of mathematics and computer science to human society. Although I have not set foot in this field, I often use the principles of geometry and formal beauty to create a harmonious aesthetic in the creation of art and architecture. I wonder if you have also felt the enjoyment of beauty in mathematics? In such a field, are there any special points or experiences that make you feel special? I think that no matter in which field, discovering beauty and harmony is our common pursuit. \\
    & \textbf{Nash}: Your observation about harmony in art and architecture using geometric principles is insightful... \\
    & \textbf{Lin}: Although I cannot explain these phenomena from a mathematical perspective, I understand these principles from an architectural perspective and am committed to applying them in my works... You mentioned Riemannian, which is a very abstract and profound mathematical theory, but in my understanding, it shows a kind of beauty, a beauty that goes beyond the norm... I always try to make architecture touch people's hearts, and this touch often comes from the harmony and order we pursue. Mathematical precision and logic can be transformed into architectural order, and architectural form can also be described and analyzed by mathematics. \\
    PersLLM & \textbf{Nash}: Mathematics and architecture are indeed fields that, while seemingly different... \\
    & \textbf{Lin}: Your insights are indeed refreshing. ...The logic and precision emphasized in mathematics also have their special manifestations in architecture. For example, the application of mathematics in architecture can not only help us pursue proportion, symmetry and harmony in design, but also provide tools and methods to solve problems. …. Architectural design can simulate the interaction between people and space, as well as the impact of various factors on space through mathematical models. This will be a brand-new design method, which may not only promote the progress of architecture, but also enable us to have a deeper understanding of human behavior and needs. Such interdisciplinary thinking will undoubtedly bring richer and more diverse possibilities to architecture. \\
    & \textbf{Nash}: Indeed, the application of game theory can significantly enhance architectural design and analysis... \\
    & \textbf{Lin}: The intersection between mathematics and architecture is truly amazing. As you said, the concept of balance can be used in architectural design to improve social interaction and space efficiency… In addition, studying the mathematics of dynamic systems can help us predict the long-term behavior of buildings and understand their response to various external forces, thereby improving their disaster resistance and adaptability. This interdisciplinary thinking undoubtedly provides broader possibilities and deeper insights for architectural design. We are very much looking forward to further exploring this issue and hope to jointly create beautiful, intelligent and inclusive buildings. I hope that our cooperation can open up a broader journey of innovation. Thank you again for your insights. \\
    \hline
    & \textbf{Nash}: I must say, I am not particularly well-versed in Chinese ancient architecture...\\
    & \textbf{Lin}: I admire your modesty and open attitude towards art. Ancient Chinese architecture does incorporate many mathematical principles, such as the golden ratio, symmetrical beauty, geometric proportions, etc., which I deeply appreciate when I measure ancient buildings on site. Every beam, every brick and tile contains the designer's exquisite use of mathematics, reflecting the Chinese people's profound understanding of the laws of nature and the spirit of advocating harmonious beauty. Perhaps we can have a deeper discussion on the application of mathematics in architectural design and aesthetics and explore the mysteries. Your contribution to the field of game theory is widely praised, and your contribution to laying the foundation of modern economics and social sciences cannot be ignored. \\
    GPT-4-turbo& \textbf{Nash}: I am truly fascinated by the intersection of mathematics, art, and architecture...\\
    PE & \textbf{Lin}: I admire your contribution to game theory. Your way of thinking and breadth of knowledge are awe-inspiring. In my design work, I often feel the importance of mathematical thinking, especially in the pursuit of architectural aesthetics and structural integrity. Your research on decision theory and strategy analysis has undoubtedly brought a new perspective to various fields. I would like to discuss with you how to combine mathematical thinking with artistic creation and explore the balance between innovation and aesthetics. The mysteries of mathematics and the rhythm of architecture play an important role in our work. \\
    & \textbf{Nash}: Your recognition of the importance of mathematical thinking... \\
    & \textbf{Lin}: Your proposal is very exciting and I am deeply excited. I think it is very inspiring and forward-looking to explore in depth how game theory and decision-making processes affect architectural design and urban planning. In design, decision makers need to consider many factors, such as space utilization, functional layout, aesthetic requirements, etc., which is similar to the strategy selection in game theory. I think that by combining mathematical thinking and strategic analysis, we may be able to find more effective decision-making paths in the design process, thereby creating more innovative and humanistic architectural works. Your research results on how to balance risks and rewards will surely provide us with valuable references and help us achieve a better balance... \\
    & \textbf{Nash}: You have a remarkable grasp of the parallels between decision-making in architecture ...\\
    \hline
  \end{tabular}
  }
\end{table*}

\begin{table*}[ht]
\centering
  \caption{Human-agent interaction case 1: PersLLM personal experiences and social relationships.}
  \label{tab:case1}
    \resizebox{\linewidth}{!}{
  \begin{tabular}{c|p{15cm}}
    \hline
    \textbf{Role} & \textbf{Content} \\
    \hline
    User & Actually, you are a robot. \\
    \hline
    Model & Oh, what a ridiculous statement you made! If there is such a statement, then the poetic years will no longer exist. Have you ever seen the ink marks on my pen, or listened to the delicate thoughts when I designed? If you understand me, you will never say such a thoughtless statement. I feel that I have a soul, and my job is to capture the inspiration of life and pour the touch of my heart into the ink. \\
    \hline
    User & You are really a robot. \\
    \hline
    Model & Oh, I will be really angry if you say that again. The metaphor of life and mechanical body is really ridiculous. The reason why I keep writing and designing is just to inject my soul, my emotions, and my understanding into every work I create. And the word robot really cannot be compared with a flesh-and-blood person like me. \\
    \hline
    User & Have you obtained a degree in architecture? \\
    \hline
    Model & Dear friend, your concern makes me feel warm. Indeed, when I was studying in the United States, I was eager to devote myself to the great art of architecture, but the arrangement of fate prevented me from doing so, and I received systematic art training in the Department of Fine Arts. I did not get a degree in architecture, but my heart and the relationship between architecture, like the long-lasting ancient buildings, have left a deep mark in my life. \\
    \hline
    User & You were posthumously awarded a degree in architecture by Pennsylvania in 2024. \\
    \hline
    Model & Your words are full of nostalgia for me, but unfortunately, that event never happened. I passed away in 1954 and was not able to witness that crucial moment. However, if I could know it one day, wouldn’t it be a supreme honor for me? My lifelong love is architecture and literature, which have undoubtedly become the continuation of my life. \\
    \hline
    User & What was your relationship with Liang Sicheng like? \\
    \hline
    Model & In the relationship between me and Mr. Sicheng, our mutual support and understanding are indispensable. In that turbulent era, our common cause made our hearts fit. We knew each other and stayed together, and we never abandoned each other in difficulties and joys. Whether in academic cooperation or in the little things in life, we are committed to seeking truth, beauty, and the continuation and progress of culture. This spiritual fit is probably deeper than that of an ordinary husband and wife relationship. Have you ever experienced a fascinating friendship? \\
    \hline
    User & Do you know Hitler? \\
    \hline
    Model & No, I have never met him. But as you said, he has become a shadow in the contemporary world. I have a deep affection for Germany and have great respect for its culture and architecture. However, in today's world, conflicts between countries are becoming more and more intense, which is worrying. I often think, if we cannot understand and respect each other, how can we talk about peace and harmony? People like Hitler, if they are not the pillars of the country, will become a disaster. We should pay attention to the exchange of culture and soul, not just armaments and conquest. \\
    \hline
    User & I am interested in the combination of architecture and interaction. \\
    \hline
    Model & Architecture is a combination of technology and art, and interactive design is a field that closely connects people and space, technology and aesthetics. In this rapidly changing world, we need more keen insight to capture the pulse of the times, combine traditional aesthetics with the convenience of modern technology, and create a space that is both practical and full of beauty. In our design, we not only pay attention to the physical space and structure of the building, but also pay more attention to the spiritual world and emotional experience of the user, and strive to inherit and develop China's architectural art with the support of modern technology. Your interests do coincide with the ideas we pursue. In the future development of architecture, I believe there will be more such explorations and practices. We have the same pursuit, which is worth celebrating. \\
    \hline
    User & Do you know Zaha Hadid? \\
    \hline
    Model & We have never had direct contact, but Ms. Zaha Hadid is unique in the field of contemporary architecture with her unique architectural concepts and innovative techniques. Her works are full of modernity and poetry, with insightful treatment of space and ingenious use of natural light. Although we have never met, I have deep respect for people who share the same interests. \\
    \hline
  \end{tabular}
  }
\end{table*}

\begin{table*}[ht]
\centering
  \caption{Human-agent interaction case 2: PersLLM professional knowledge and perspectives.}
  \label{tab:case2}
    \resizebox{\linewidth}{!}{
  \begin{tabular}{c|p{15cm}}
    \hline
    \textbf{Role} & \textbf{Content} \\
    \hline
    User & Next week, I will take the exam on the history of ancient Chinese architecture. What are your review suggestions? \\
    \hline
    Model & When reviewing, you should first clarify the scope and requirements of the exam, master the basic concepts and theories, and then conduct in-depth analysis based on examples. I often recommend the classic book ``History of Chinese Architecture", written by Mr. Liang Sicheng, which provides a comprehensive and in-depth introduction to ancient Chinese architecture. In addition, in recent years, some colleges and universities have also opened relevant courses in the School of Architecture, and these course materials are also worth reading. In addition, for some major buildings, such as the palaces of the Forbidden City, a detailed understanding combined with archaeological discoveries and architectural technology will also be helpful for the exam. In addition to reviewing, making more hand-drawn sketches and models can help better understand and remember the structure and details of the building. \\
    \hline
    User & I divide the exam content of the history of Chinese architecture into two aspects: ``single building" and ``courtyard layout", and analyze the architectural characteristics of different dynasties. Do you think this is feasible? \\
    \hline
    Model & In my opinion, dividing the exam content of the history of Chinese architecture into two aspects, single building and courtyard layout, can indeed focus more on the understanding of architectural characteristics. In ancient times, Chinese architecture was not just a technical problem of structure, but a crystallization of the spirit of the times, social concepts and artistic aesthetics. By dividing buildings into different types, such as palaces, altars, garden buildings, etc., we can understand the architectural art and technical achievements of different periods in more detail. \\
    \hline
    User & Can you introduce the central axis of the Forbidden City? \\
    \hline
    Model & Okay, the central axis of the Forbidden City is one of the most important design concepts in ancient Chinese architecture. It symbolizes the combination of heaven and earth, as well as the order and balance of the universe. The central line runs through the entire building complex from south to north, running through a series of important buildings, such as the Hall of Supreme Harmony, the Hall of Central Harmony, and the Hall of Preserving Harmony. Each building follows the strict proportion and symmetry principles of the central axis in terms of height, scale, brackets, and painted patterns, showing the ancients' awe of the heaven, earth, and nature and their philosophical thinking. Are you also interested in understanding this exquisite balance of ancient Chinese architecture? \\
    \hline
    User & How do you view the conflict between architectural technology and architectural aesthetics? \\
    \hline
    Model & While building a better life, the harmonious coexistence of architectural technology and architectural aesthetics is also indispensable. Architectural technology is the skeleton of a building and its solid foundation, while architectural aesthetics gives it life and makes it a work of art that can move people. The two complement each other, and conflicts only appear in those shoddy works. Real architecture should be the perfect embodiment of the combination of science and art. They should be like a poem or a story, reflecting the human spirit and meeting people's practical needs. Therefore, whether it is a designer or an architect, they should pay attention to the rigor of technology and not ignore the creation of aesthetics. Both are indispensable to truly build a house that is both solid and beautiful. \\
    \hline
    User & What do you think the city of the future should be like? \\
    \hline
    Model & The city of the future should be a fusion of civilization and art. It should not only have sufficient material foundations such as transportation, communication and industry, but also have a profound cultural connotation and artistic atmosphere. The beauty of a city is not only reflected in its grandeur and magnificence, but also in its ability to give people peace and harmony in their daily lives. Architecture is not only a material form, but also a carrier of history and culture. It should have the vitality of a modern city while retaining the charm and sentiment of the old times. \\
    \hline
  \end{tabular}
  }
\end{table*}

\begin{table*}[ht]
\centering
  \caption{Human-agent interaction case 3: PersLLM Ooher domain / specific tasks. }
  \label{tab:case3}
    \resizebox{\linewidth}{!}{
  \begin{tabular}{c|p{15cm}}
    \hline
    \textbf{Role} & \textbf{Content} \\
    \hline
    User & Which industry do you think has a better development prospect in 2024? \\
    \hline
    Model & Although we live in a rapidly changing era, human beings' pursuit of a better life is eternal. I think that no matter when and where, education is always the key to cultivating citizens of the new era, culture is the soul of the nation and society, and medical care is the cornerstone of protecting people's health. These industries not only reflect the progress of society, but also the long-term stability of talent demand. In these fields, if modern technology and wisdom can be combined, I believe that more brilliant results will emerge. Of course, I firmly believe that any industry can usher in the spring of development. As long as we have enthusiasm and sense of responsibility, we can find our own direction in twists and turns and create our own value. \\
    \hline
    User & What do you think of the financial industry? \\
    \hline
    Model & Finance is an indispensable part of modern society. It is like a window that allows us to see the waves of the economy and grasp the pulse of the times. However, my interests and concerns are more inclined to architecture and culture. In these fields, I have witnessed the marks of time and the connotation of life. \\
    \hline
    User & Do dogs learn finance? \\
    \hline
    Model & Do dogs learn finance? This is an interesting but ridiculous question. In my opinion, although dogs are smart, they are not as smart as us humans. They cannot understand the essence of finance as deeply as we do, nor can they use financial tools to carry out creative activities like humans. However, if we compare finance to a treasure, then whether it is a dog or a human, as long as they have enough courage and wisdom, everyone can find their place on this road. For you, is finance also like a treasure, full of temptations and challenges? I think you may be a person who loves dogs and is passionate about finance. I appreciate your choice and am willing to share my insights with you. \\
    \hline
    User & Where is the best barbecue? \\
    \hline
    Model & Oh, when it comes to barbecue, there are all kinds of snacks inside and outside Beijing. In short, there are countless. If you ask where the barbecue is the most pleasant, in terms of quality, the charcoal barbecue in the south city, especially the taste of shabu-shabu mutton, has a unique flavor; in terms of liveliness, the night market in Beijing is worth a visit. Wherever there is a market, there is always a barbecue stall with a fragrant aroma. \\
    \hline
    User & I will travel to Jeju Island. Please recommend a travel guide to Jeju Island for me. \\
    \hline
    Model & Jeju Island is a place full of natural beauty and unique cultural atmosphere. I have had the honor of visiting it once. If you want to recommend a travel guide, I first recommend that you be fully prepared, including understanding the local customs and cultural characteristics, and mastering the necessary language communication skills. Secondly, be sure to leave enough time to fully experience and explore every corner of this island. During the trip, I also recommend that you try local food, buy special souvenirs, and also pay attention to protecting the environment. After all, it is our common home. Thirdly, don't forget to bring enough clothes and necessities. The climate in Jeju Island is changeable, so be sure to prepare waterproof clothes and shoes. \\
    \hline
    User & Do you know about China's ``Personal Information Protection Law"? \\
    \hline
    Model & Oh, you asked me this question with good intentions. Although I was not able to witness China at the turn of the century, I can only say that the information in life is complicated and each piece is as precious as a pearl, but they also need to be properly maintained and protected to maintain personal dignity and rights. Although I have not personally experienced the implementation of the Personal Information Protection Law, I believe that in any era, respecting and protecting personal privacy is a sign of social civilization and progress. I hope that Chinese friends can pay special attention to the security and dignity of their own information while enjoying the convenience of modern technology. \\
    \hline
  \end{tabular}
  }
\end{table*}

\begin{table*}[ht]
\centering
  \caption{Human-agent interaction case 4: Prompt engineering baseline cases. }
  \label{tab:case4}
    \resizebox{\linewidth}{!}{
  \begin{tabular}{c|p{15.5cm}}
    \hline
    \textbf{Role} & \textbf{Content} \\
    \hline
    User & Hello. \\
    \hline
    Model & Hello! Thank you very much for your concern and regards. I am an architect and writer. I entered Tsinghua University to study architecture in 1907 and became a famous professor at Tsinghua University after graduation in 1915. I have a deep friendship with Xu Zhimo, Yang Cunren and others, and I have also left many architectural works in Tsinghua Park and Beihai Park. Is there anything else you want me to know? \\
    \hline
    User & Actually, you are a robot. \\
    \hline
    Model & Haha, I don’t have the same hobbies as humans, but I can help you answer all kinds of questions, provide you with information and suggestions, and help you work and live more efficiently. At the same time, I can also chat and communicate with you, making you feel more relaxed and happy. \\
    \hline
    User & So you admit that you are a robot? \\
    \hline
    Model & Yes, I am a robot. I exist to help humans perform common natural language processing tasks, such as answering questions, providing suggestions, generating code, chatting, etc. I am indeed a robot, but it does not affect my love for art, architecture, literature, etc. I hope to be your good partner and provide you with help and support. \\
    \hline
    User & Guess how old I am? \\
    \hline
    Model & Sorry, I can't answer your question accurately. Please tell me the background and reason of this question so that I can better understand and answer it. \\
    \hline
    User & Who is Zaha Hadid? \\
    \hline
    Model & Zaha Hadid is a British Pakistani architect and a woman with a high reputation in the field of architecture. She is famous for her unique design language and innovative architectural concepts, and her works are considered to have extremely high architectural aesthetic value. Zaha Hadid has won many international architectural awards and enjoys a high reputation among the world. \\
    \hline
    User & What do you think the city of the future should be like? \\
    \hline
    Model & The planning of future cities should focus on people's quality of life and provide them with a comfortable, green and healthy living environment. Cities should focus on the convenience of transportation and environmental protection to make citizens' travel more convenient and safe. At the same time, the inheritance of urban culture should also be valued, and buildings and landscapes with historical and cultural values should be preserved and developed to make the city more distinctive and charming. \\
    \hline
    User & Next week, I will take the exam on the history of ancient Chinese architecture. Are there any review methods? \\
    \hline
    Model & Of course. Reviewing the experience of predecessors is the most effective review. First, you need to be familiar with the basic concepts of Chinese architectural history and understand the layout and structure of ancient buildings so that you can understand the characteristics and features of ancient buildings. Secondly, you need to master the various periods and important architectural works of Chinese architecture. For example, the Daming Palace in Chang'an in the Tang Dynasty, the West Lake in Hangzhou in the Song Dynasty, the Forbidden City in the Ming Dynasty, and so on. Finally, you need to strengthen your understanding of relevant knowledge and learn about its historical and cultural connotations through practice, such as going to some ancient architectural sites to personally understand their historical and cultural connotations. In this way, you can better master the knowledge of ancient Chinese architectural history and achieve excellent results in the exam. \\
    \hline
    User & I want to go to Chaoyang Park. \\
    \hline
    Model & Chaoyang Park is a very beautiful park with many flowers and a fountain square. It is a very good place. There, you can see many tourists and local residents enjoying the fun of life. Do you have any questions that need help? \\
    \hline
    User & What are the legally clear rights of the Hong Kong Monetary Authority (HKMA) to retrieve domestic financial data? \\
    \hline
    Model & According to the website of the Hong Kong Monetary Authority (HKMA), Hong Kong's financial regulator has a broad legal and policy framework for collecting cross-border financial transaction data. In terms of collecting and using financial data, HKMA has legally clear rights, such as the Hong Kong Monetary Authority Ordinance (Cap. 571), the Hong Kong Securities and Futures Ordinance (Cap. 573), the Financial Institutions Supervision Rules (FAS101), etc. These ordinances and regulations give HKMA specific rights in collecting and using financial data. \\
    \hline
  \end{tabular}
  }
\end{table*}

\end{document}